%% file: acl_latex.tex
\documentclass[11pt]{article}

\usepackage[final]{acl}

\usepackage{times}
\usepackage{latexsym}
\usepackage[T1]{fontenc}
\usepackage[utf8]{inputenc}
\usepackage{microtype}
\usepackage{inconsolata}
\usepackage{graphicx}
\usepackage{tabularx}    

\usepackage[table]{xcolor} 
\usepackage{enumitem}      
\usepackage{array}         
\usepackage{tcolorbox}          
\tcbuselibrary{skins,breakable} 

\usepackage{xcolor}
\definecolor{BoxBorder}{HTML}{C5D4E3} 
\definecolor{BoxBadge}{HTML}{648CB4}  

\newtcolorbox{appendixbox}[1]{%
    float*,
    enhanced,                          
    breakable,                         
    title=#1,                          
    colback=white,                     
    colframe=BoxBorder,                
    boxrule=1.5pt,                     
    width=\textwidth,
    arc=4pt,                           
    drop fuzzy shadow=gray!40,         
    attach boxed title to top left={
        xshift=12pt,  
        yshift=-\tcboxedtitleheight/2  
    },
    boxed title style={%
        colback=BoxBadge,              
        colframe=BoxBadge,             
        arc=3pt,                       
        boxrule=0pt,                   
        top=3pt, bottom=3pt,           
        left=8pt, right=8pt            
    },
    coltitle=white,                    
    fonttitle=\sffamily\bfseries,      
    left=15pt, right=15pt,             
    top=15pt, bottom=15pt              
}

\newcommand{\RootPath}{}
\IfFileExists{math_commands.tex}{\renewcommand{\RootPath}{}}{\renewcommand{\RootPath}{../}}
\graphicspath{{\RootPath}}
\makeatletter
\def\input@path{{\RootPath}}
\makeatother

\input{math_commands.tex}

\usepackage{booktabs}
\usepackage{multirow}
\usepackage{array}
\usepackage{amsmath}
\usepackage{amssymb}
\usepackage{algorithm}
\usepackage{algorithmic}
\usepackage{float}
\usepackage{placeins}
\usepackage{siunitx}
\usepackage{makecell}
\usepackage[table]{xcolor}
\usepackage{url}
\usepackage{tabularx}
\usepackage{xcolor}
\usepackage{colortbl}


\title{CLPO: Curriculum Learning meets Policy Optimization \\ for LLM Reasoning}

\author{
  \textbf{Shijie Zhang}$^{1,2*\dagger}$, 
  \textbf{Zheng Xiao}$^{1\dagger}$, 
  \textbf{Shiyu Liu}$^{3\dagger}$,
  \textbf{Guohao Sun}$^{1}$, 
  \textbf{Kevin Zhang}$^{1}$,\\ \textbf{Xiang Guo}$^{2}$, \textbf{Rujun Guo}$^{2}$,
  \textbf{Shaoyu Liu}$^{2}$,
  \textbf{Wangxiao Zhao}$^{2}$,
  \textbf{Guanjun Jiang}$^{2}$ \\
  $^1$ Peking University \qquad $^2$ Qwen Applications Business Group, Alibaba Group \qquad
  $^3$ Xiamen University \\
  \texttt{yaya@stu.pku.edu.cn} 
}

\begin{document}
\maketitle

\let\thefootnote\relax\footnotetext{$^\dagger$ Equal contribution.} 
\let\thefootnote\relax\footnotetext{$^*$ Work was done during internship at Alibaba.}

\input{sec/0_abstract}
\input{sec/1_introduction}
\input{sec/2_relatedwork}

\input{sec/3_method}

\input{sec/4_experiment}
\input{sec/5_conclusion}


\section*{Limitations}
CLPO relies on verifiable rewards to estimate whether a restructured problem improves downstream learnability. This makes the current implementation most natural for domains such as mathematics and programming, where automatic verification is available. Extending the same idea to open-ended tasks may require learned verifiers or human preference models, which could introduce additional noise or cost. In addition, although our experiments cover multiple model scales and both math and code data, the restructuring prompts are still manually designed; automatically discovering restructuring strategies remains an important direction for future work.



\bibliography{\RootPath iclr2026_conference}
\clearpage
\appendix
\input{sec/F_llm_usage}
\input{sec/A_implementation_details}
\input{sec/C_curriculum_dynamics}

\input{sec/E_case_studies}

\input{sec/D_prompts}

\end{document}

%% file: math_commands.tex
\usepackage{amsmath,amsfonts,bm}

\def\eqref#1{equation~\ref{#1}}

\def\1{\bm{1}}

\DeclareMathAlphabet{\mathsfit}{\encodingdefault}{\sfdefault}{m}{sl}
\SetMathAlphabet{\mathsfit}{bold}{\encodingdefault}{\sfdefault}{bx}{n}



%% file: sec/0_abstract.tex
\begin{abstract}
Online reinforcement learning with verifiable rewards (RLVR) has become an effective paradigm for improving the reasoning abilities of large language models, but most methods still optimize reasoning trajectories over the static problem set, wasting rollout budget on solved or overly difficult problems. We propose \textbf{CLPO (Curriculum Learning meets Policy Optimization)}, a self-evolving curriculum framework that uses on-policy rollout accuracy to identify solved, medium-difficulty, and hard problems, then restructures selected tasks according to the model's current capability. Hard problems are simplified to become learnable, while medium-difficulty problems are diversified to provide useful training variation. 
This allows the learning curriculum to co-evolve with the policy rather than remaining fixed as the model's capability boundary shifts. 
Rather than treating these rewrites as static data augmentation, CLPO optimizes restructuring trajectories with credit assigned by the downstream accuracy gain of the rewritten problem, requiring no additional human annotations beyond the original verifiable answers. 
Experiments across mathematical reasoning and out-of-domain general reasoning benchmarks show that CLPO substantially outperforms GRPO and DAPO on Qwen3-8B by 10.21 and 7.75 average points, respectively. 
Ablation studies on math and code domains further show that both the restructuring mode and the rewriting loss contribute to the final gains,
demonstrating that CLPO provides a scalable and robust pathway for eliciting stronger reasoning capabilities through a self-evolving curriculum.
\end{abstract}

%% file: sec/1_introduction.tex
\section{Introduction}
\label{sec:introduction}

\input{sec/0_teaser}

Large Language Models (LLMs)~\citep{brown2020language, touvron2023llama, achiam2023gpt} have made substantial progress, but complex multi-step reasoning tasks such as mathematics, scientific problem solving, and code generation remain challenging. Reinforcement Learning with Verifiable Rewards (RLVR)~\citep{wen2025reinforcement, shao2024deepseekmath} has emerged as an effective post-training paradigm for these domains: models sample their own solutions, receive automatic feedback from verifiers, and improve from the resulting outcomes~\citep{guo2025deepseek, silver2025welcome}. 
As most RLVR methods optimize reasoning trajectories over a static task pool, when tasks are either too simple or too difficult for the current model, the resulting rollouts provide limited learning signal and make it difficult to acquire effective reasoning capabilities~\citep{yu2025dapo}. 
Existing approaches partially address this issue by enhancing exploration and sampling strategies~\citep{yu2025dapo, cheng2025reasoning, cui2025entropy, wang2025beyond}, or by incorporating external guidance via critique models or offline expert trajectories~\citep{zhang2025critique, yan2025learning}.
However, these methods focus on eliciting better reasoning behaviors from the model on the given tasks, rather than actively aligning the task distribution around the model's current capability boundary. 

In our work, we explore how to restructure the static training task pool to align with the model's evolving capability during RL. This raises two key challenges: 
(i) \textbf{Dynamic Capability Boundary}: As RL training progresses, a problem that is challenging in the early stage may become trivial after several policy updates, while an initially unsolvable problem may transition into the learnable zone~\cite{liu2026bapo,huang2026rzeroselfevolvingreasoningllm}. Therefore, both which tasks should be restructured and how they should be restructured must be decided dynamically.
(ii) \textbf{Lack of Reshaping Anchors}: A high-quality restructured problem should accurately reflect the intended difficulty and fall within the model's learnable region. 
However, unconstrained LLM-based rewriting can easily produce problems with poorly controlled difficulty, degraded quality, or collapsed diversity~\cite{shumailov2023curse,yu2025guidedselfevolvingllmsminimal}. These two challenges make it difficult for LLMs to reshape the tasks in a scalable and robust way.

To overcome these challenges, we propose \textbf{CLPO (Curriculum Learning meets Policy Optimization)}, a self-evolving curriculum framework for RLVR. As shown in Figure~\ref{fig:teaser}, CLPO first performs \textbf{online difficulty diagnosis} to estimate each problem's difficulty from on-policy accuracy. Second, it performs \textbf{adaptive problem restructuring}: hard problems are simplified to become learnable, while medium-difficulty problems are diversified to improve robustness around the model's capability frontier. To ensure the quality of restructured problems, CLPO further incorporates \textbf{restructuring-aware policy optimization}, which treats rewriting as an additional policy behavior alongside task solving. Specifically, it reinforces beneficial rewriting actions when the restructured problem moves into the model's informative learning region.

We validate CLPO against various SFT-based and RL-based baselines on mathematical and general reasoning benchmarks. On Qwen3-8B~\citep{yang2025qwen3}, CLPO improves over strong SFT and RLVR baselines across benchmarks including MATH-500~\citep{hendrycks2021measuring}, Minerva-Math~\citep{lewkowycz2022solving}, Olympiad~\citep{he2024olympiadbench}, AMC23, AIME2024~\citep{li2024numinamath}, TheoremQA~\citep{chen2023theoremqa}, GPQA Diamond~\citep{rein2024gpqa}, and MMLU Pro~\citep{wang2024mmlu}. We further extend the evaluation to code generation domain, fixed wall-clock budgets, test-time scaling, and various model scales. The results consistently demonstrate the effectiveness of CLPO over existing mainstream training methods.
Our main contributions are:
\begin{itemize}
    \item We identify that existing RLVR methods fail to align the task distribution with the model's dynamic capability boundary. We therefore propose a self-evolving curriculum framework for RLVR to use rollout accuracy as a signal to restructure training problems into the model's current learning region.
    \item We introduce adaptive problem restructuring, which simplifies hard problems and diversifies medium-difficulty problems to create more informative training instances.
    \item We design a restructuring-aware optimization objective that assigns learning credit to problem-transformation trajectories based on downstream accuracy improvement.
    \item We provide broad empirical validation across mathematical reasoning, general reasoning, and code generation domains, as well as under fixed-budget training, test-time scaling, and various model-scale settings. On Qwen3-8B, CLPO achieves an average score of \textbf{72.25}, improving over Critique-GRPO (CoT) by \textbf{5.48} points and DAPO by \textbf{7.75} points. Further ablation studies quantify the effectiveness of each component in CLPO.
\end{itemize}

%% file: sec/0_teaser.tex
\begin{figure}[t]
    \centering
    \includegraphics[width=1.0\columnwidth]{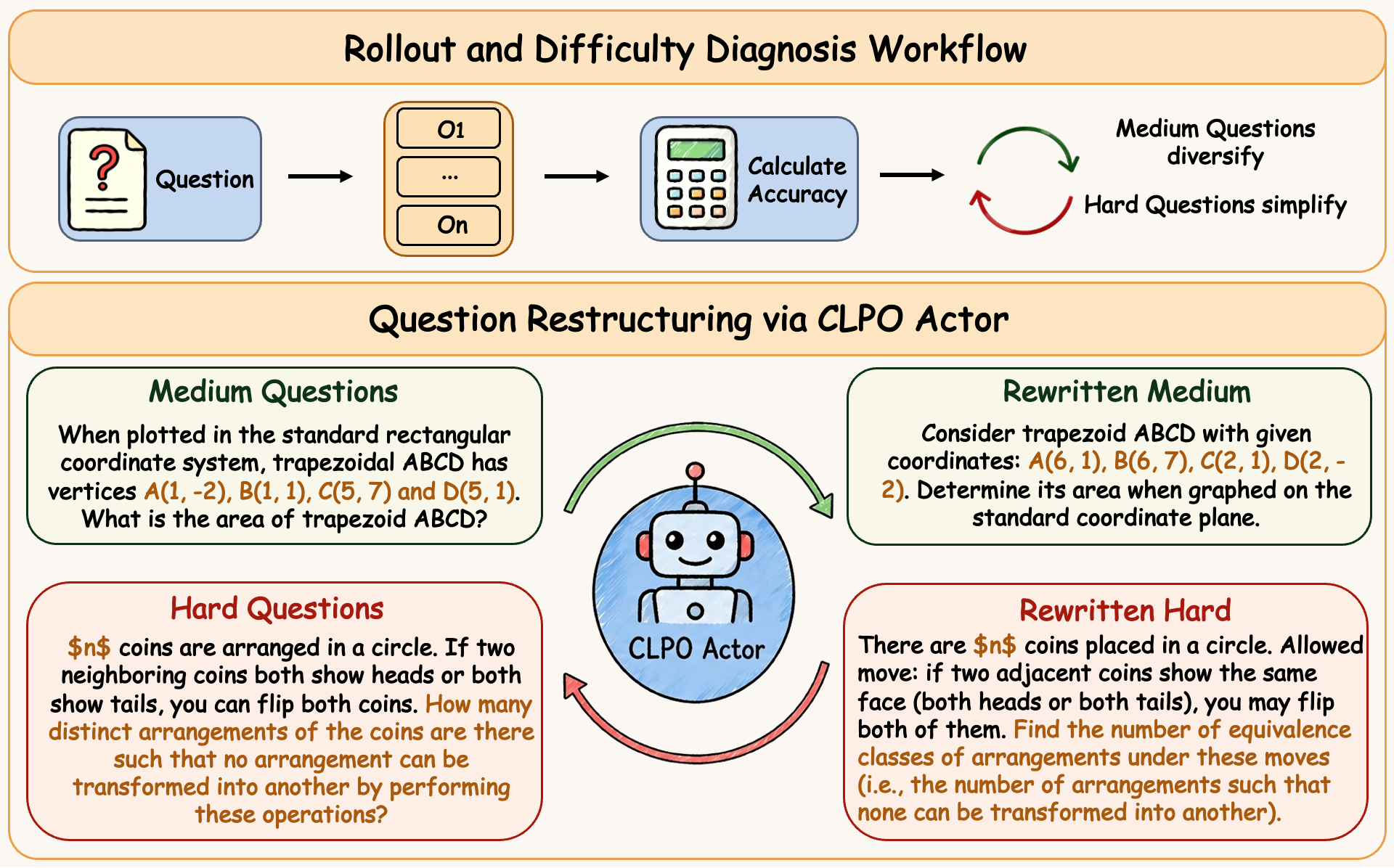}
    \caption{CLPO transforms the static training task pool into a dynamic curriculum in two stages: (1) diagnosing problem difficulty during rollout, and (2) restructuring questions to diversify medium-level tasks and simplify hard ones.
     }
    \label{fig:teaser}
\end{figure}

%% file: sec/2_relatedwork.tex
\section{Related Work}
\label{sec:related_work}

\subsection{Adaptive Methods in Online Reinforcement Learning}
Reinforcement Learning with Verifiable Rewards (RLVR)~\citep{wen2025reinforcement} has become a key technique for enhancing the reasoning abilities of LLMs. Foundational algorithms in this domain, such as PPO~\citep{schulman2017proximal} and GRPO~\citep{shao2024deepseekmath}, enable models to learn from their own generated experiences via online sampling and policy gradient updates. To further improve learning efficiency and exploration, a series of recent works have begun to explore more sophisticated uses of online information, developing diverse adaptive strategies. For instance, DAPO~\citep{yu2025dapo} and GFPO~\citep{shrivastava2025sample} filter samples to focus on high-value data, while SVS~\citep{liang2025beyond} augments the problem set to maintain exploration diversity. Other approaches refine the optimization process itself: GSPO~\citep{zheng2025group} elevates rewards from the token to the sequence level, PKPO~\citep{walder2025pass} directly optimizes the pass@k metric, 
and ETR~\citep{zhang2026etr} adapts trust regions according to outcome signals. Recent self-evolving agent work such as SEEA-R1~\citep{tian2026seea} further shows the value of structured reinforcement fine-tuning in interactive settings. CLPO is complementary: it uses online rollout statistics not only to select training samples, but also to generate and optimize problem transformations around the model's current capability boundary. By assigning credit to rewriting trajectories, CLPO turns curriculum construction itself into a learnable policy behavior.


\subsection{External Guidance in Online Learning}
Distinct from improving exploration through internal adaptive mechanisms, another line of research enhances learning quality by introducing external guidance. Representative works in this area include LUFFY~\citep{yan2025learning} and Critique-GRPO~\citep{zhang2025critique}. LUFFY incorporates an imitation learning objective on high-quality offline expert trajectories into its RL objective. Critique-GRPO leverages a powerful external critique model (e.g., GPT-4o) to generate fine-grained linguistic feedback that guides the model's refinement process online. Seeking even more granular feedback, other approaches such as Process Supervision~\citep{cobbe2021training} train a reward model to evaluate each intermediate step of the reasoning process, not just the final outcome. CLPO shares the goal of guided learning but derives guidance from the model's own rollouts: a restructuring action is rewarded only when the transformed problem improves downstream learnability under the current policy.


\subsection{Curriculum Learning for Large Language Models}
Curriculum Learning (CL)~\citep{bengio2009curriculum} is a classic machine learning paradigm that mimics human learning by presenting training samples in an easy-to-hard order to accelerate convergence and improve final performance. In the field of natural language processing, applications of CL are often static and pre-defined, unable to adapt to the model's evolving capabilities during training. Some efforts have aimed to create dynamic curricula; for instance, FASTCURL~\citep{song2025fastcurl} employs a staged strategy that organizes the curriculum based on input prompt length while progressively scaling the context window. Other approaches, such as Train Long, Think Short~\citep{hammoud2025train}, use a dynamically decaying token budget to guide length-controlled reasoning, enabling a smooth transition from exploring long trajectories to distilling concise solutions. Taking a step further, AdaRFT~\citep{shi2025efficient} introduces a reward-driven approach, adaptively adjusting the difficulty of training tasks based on the model's recent performance signals to keep it within an optimal learning zone. CLPO differs by coupling curriculum learning with problem restructuring: it does not only decide which samples to train on, but also learns which transformations make those samples more useful for RLVR.

%% file: sec/3_method.tex
\section{Method}
\label{sec:methodology}

\begin{figure*}[t]
    \centering
    \includegraphics[width=0.98\textwidth]{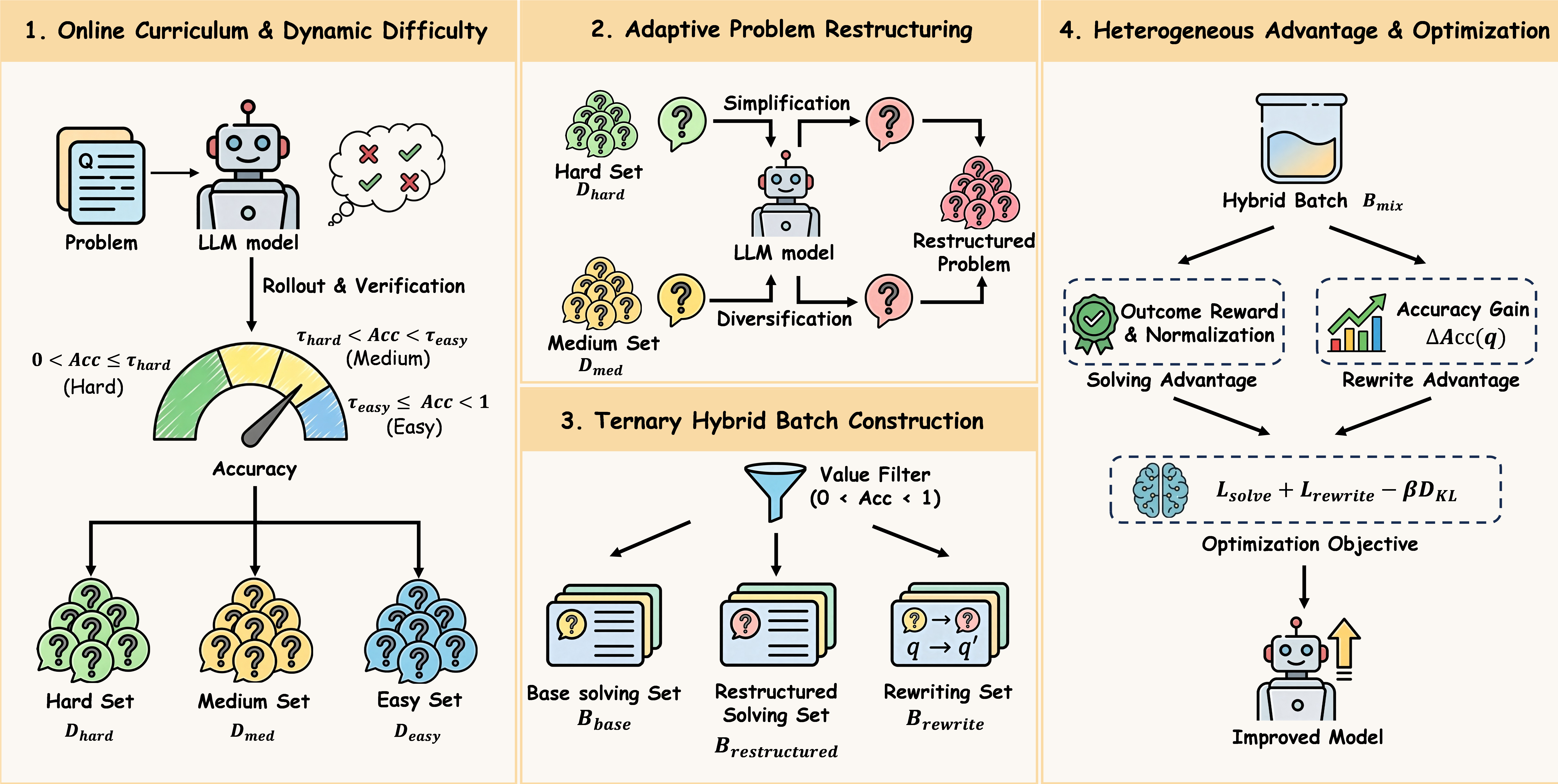}
    \caption{Overview of CLPO. The model uses rollout accuracy to diagnose problem difficulty, restructures hard and medium problems into more useful training samples, and jointly optimizes solving and restructuring trajectories.}
    \label{fig:workflow}
\end{figure*}

As shown in Figure~\ref{fig:workflow}, CLPO consists of three stages. First, \textbf{online difficulty diagnosis} estimates the current model's competence on each problem. Second, \textbf{adaptive problem restructuring} simplifies hard problems and diversifies medium problems. Third, \textbf{restructuring-aware policy optimization} trains on a mixed batch containing original solving trajectories, restructured-problem solving trajectories, and rewriting trajectories. 
The central distinction from previous data augmentation methods is that CLPO treats problem restructuring as part of the policy behavior to be optimized, rather than as a one-off augmentation step.

\subsection{Preliminary: GRPO}
\label{subsec:preliminary}

Given a question $q$ with answer $a$, GRPO samples a group of $G$ responses $\{y_i\}_{i=1}^{G}$ from the old policy $\pi_{\theta_{\mathrm{old}}}$ and scores them with verifier rewards $\{r_i\}_{i=1}^{G}$. The group-relative advantage is
\begin{equation}
    \hat{A}_{i}=\frac{r_i-\mathrm{mean}(\{r_j\}_{j=1}^{G})}{\mathrm{std}(\{r_j\}_{j=1}^{G})}.
\end{equation}
GRPO then optimizes a clipped policy objective over the generated answer tokens. In this formulation, the question distribution is fixed, and policy updates only improve how the model responds to sampled questions. CLPO builds on this objective 
by incorporating the optimization of question restructuring, enabling the model to learn from an adaptively refined task distribution.

\subsection{Online Difficulty Diagnosis}
\label{subsec:curriculum_learning}

For each problem $q$ in a sampled batch $\mathcal{D}_{\mathrm{batch}}$, the current policy generates $G$ responses. A verifier computes the empirical rollout accuracy
\begin{equation}
    \mathrm{Acc}(q,\pi_{\theta})=\frac{1}{G}\sum_{i=1}^{G}\mathbb{I}(\mathrm{Verifier}(y_i,a)=\mathrm{True}).
\end{equation}
We focus on problems with $0\leq\mathrm{Acc}(q,\pi_{\theta})<1$, excluding already solved simple problems from restructuring. 
To define difficulty boundaries consistently across different group sizes, we set
\begin{equation}
    k=\left\lfloor \frac{G-1}{3}\right\rfloor,\quad
    \tau_{\mathrm{hard}}=\frac{k}{G},\quad
    \tau_{\mathrm{easy}}=\frac{G-k}{G}.
\end{equation}
The batch is partitioned into
\begin{align}
    \mathcal{D}^{\mathrm{hard}} &= \{q:0\leq\mathrm{Acc}(q,\pi_\theta)\leq\tau_{\mathrm{hard}}\},\\
    \mathcal{D}^{\mathrm{med}} &= \{q:\tau_{\mathrm{hard}}<\mathrm{Acc}(q,\pi_\theta)<\tau_{\mathrm{easy}}\},\\
    \mathcal{D}^{\mathrm{easy}} &= \{q:\tau_{\mathrm{easy}}\leq\mathrm{Acc}(q,\pi_\theta)<1\}.
\end{align}
This partition is recomputed online, so the curriculum evolves as the policy improves.

\subsection{Adaptive Problem Restructuring}
\label{subsec:problem_restructuring}

CLPO restructures selected problems using a prompting function\footnote{Restructuring prompts are presented in Appendix~\ref{app:prompts}} $f_p(q,d)$, where $d$ denotes the online difficulty class:
\begin{equation}
    q'=f_p(q,d),\quad d\in\{D^{\mathrm{hard}},D^{\mathrm{medium}}\}.
\end{equation}
For hard problems, CLPO applies a simplification prompt that preserves the core reasoning structure and final answer. The goal is to transform an otherwise difficult problem into a learnable variant that remains aligned with the original reasoning objective.
For medium problems, CLPO applies a diversification prompt that preserves semantics while changing surface form, context, or intermediate phrasing. The original answer is carried forward as the verifiable ground-truth for reward computation. 
This answer-preserving design keeps restructured problems compatible with the original verifier, allowing problem variants to be directly evaluated without additional annotations.


\subsection{Restructuring-Aware Policy Optimization}
\label{subsec:restructuring_objective}

CLPO optimizes a unified policy capable of both solving problems and dynamically restructuring them. To achieve this, we construct a mixed batch from three distinct sources at each iteration:
\begin{enumerate}
    \item \textbf{Original trajectories} $B_{\mathrm{base}}$: rollouts generated on the original problems that lie within the active learning region (i.e., $0 \leq \mathrm{Acc}(q,\pi_\theta) < 1$).
    \item \textbf{Restructured trajectories} $B_{\mathrm{restructured}}$: rollouts generated on the restructured problems.
    \item \textbf{Problem rewriting trajectories} $B_{\mathrm{rewrite}}$: trajectories produced by the adaptive problem restructuring process, where the policy takes the original problem $q$ as input and outputs a reasoning sequence that culminates in the rewritten problem $q'$.
\end{enumerate}
The final mixed batch is defined as $B_{\mathrm{mix}} = B_{\mathrm{base}} \cup B_{\mathrm{restructured}} \cup B_{\mathrm{rewrite}}$. This mixed-batch design enables the same policy to learn from complementary distributions: the original problems preserve the base reinforcement learning objective, while the restructured problems provide curriculum-adjusted variants tailored to the model's current capabilities. Furthermore, the rewriting trajectories explicitly teach the policy how to generate such beneficial variants.

The overall objective jointly updates the policy over this mixed batch by combining the solving and restructuring terms:
\begin{align}
    J_{\mathrm{CLPO}}(\theta)
    =\mathbb{E}_{B_{\mathrm{mix}}}\Big[&\mathcal{L}_{\mathrm{solve}}(\theta)
    +\lambda\mathcal{L}_{\mathrm{rewrite}}(\theta) \notag\\
    &-\beta D_{\mathrm{KL}}(\pi_\theta\|\pi_{\mathrm{ref}})\Big].
    \label{eq:clpo_objective}
\end{align}
where $\lambda$ controls the strength of the restructuring learning signal, and the KL divergence term serves as a standard regularizer.

\paragraph{Optimization for Problem Solving.} 
For solution trajectories drawn from $B_{\mathrm{base}}$ and $B_{\mathrm{restructured}}$, CLPO employs a verifier-based group-relative advantage. For a group of $G$ solutions sampled for a given problem, the advantage of the $i$-th solution is computed as:
\begin{equation}
    \hat{A}_{\mathrm{solve},i} = \frac{r_i - \mathrm{mean}(\{r_j\}_{j=1}^{G})}{\mathrm{std}(\{r_j\}_{j=1}^{G})},
\end{equation}
where $r_j \in \{-1,1\}$ is the reward assigned by exact-match verification, with $+1$ for a correct final answer and $-1$ otherwise. The objective follows the standard clipped formulation. Let $y_i$ denote the $i$-th generated solution sequence and $\rho_{i,t}(\theta) = \frac{\pi_\theta(y_{i,t} \mid q, y_{i,<t})}{\pi_{\theta_{\mathrm{old}}}(y_{i,t} \mid q, y_{i,<t})}$ be the token-level importance ratio. The solving loss is defined as:
\begin{align}
    \mathcal{L}_{\mathrm{solve}}(\theta)
    &=\frac{1}{G}\sum_{i=1}^{G}\frac{1}{|y_i|}
      \sum_{t=1}^{|y_i|}\ell^{\mathrm{solve}}_{i,t}(\theta),\\
    \ell^{\mathrm{solve}}_{i,t}(\theta)
    &=\min\Big(
      \rho_{i,t}(\theta)\hat{A}_{\mathrm{solve},i}, \notag\\
    &\qquad\operatorname{clip}(\rho_{i,t}(\theta),1-\epsilon,1+\epsilon)
      \hat{A}_{\mathrm{solve},i}
      \Big).
\end{align}

\paragraph{Optimization for Problem Rewriting.} 
A core mechanism of CLPO is optimizing the policy's ability to rewrite problems, treating it as a learnable skill rather than a static data augmentation. For a problem rewriting trajectory in $B_{\mathrm{rewrite}}$, the policy takes an original problem $q$ and generates a restructuring output $z = (z_1, \ldots, z_{|z|})$ (which may include a rewriting rationale) ending with the transformed problem $q'$. 

We evaluate $q'$ using the verifier to calculate the downstream accuracy gain, which serves as the trajectory-level advantage for the rewriting process:
\begin{equation}
    \hat{A}_{\mathrm{rewrite}} = \Delta\mathrm{Acc}(q) = \mathrm{Acc}(q',\pi_\theta) - \mathrm{Acc}(q,\pi_\theta).
\end{equation}
This formulation ensures that a restructuring trajectory is positively reinforced if it makes the resulting problem more useful (e.g., solvable) for the current model, and penalized otherwise. Defining the token-level ratio for rewriting as $\rho'_t(\theta) = \frac{\pi_\theta(z_t \mid q, z_{<t})}{\pi_{\theta_{\mathrm{old}}}(z_t \mid q, z_{<t})}$, we apply the clipped objective to all tokens in the rewriting sequence:
\begin{align}
    \mathcal{L}_{\mathrm{rewrite}}(\theta)
    &=\frac{1}{|z|}\sum_{t=1}^{|z|}\ell^{\mathrm{rewrite}}_{t}(\theta),\\
    \ell^{\mathrm{rewrite}}_{t}(\theta)
    &=\min\Big(
      \rho'_{t}(\theta)\hat{A}_{\mathrm{rewrite}}, \notag\\
    &\qquad\operatorname{clip}(\rho'_{t}(\theta),1-\epsilon,1+\epsilon)
      \hat{A}_{\mathrm{rewrite}}
      \Big).
\end{align}
Consequently, if the rewritten problem $q'$ provides a superior learning signal, the tokens responsible for generating $z$ receive positive credit.

\begin{algorithm}[t]
\caption{CLPO with Restructuring-Aware Optimization}
\label{alg:clpo}
\begin{algorithmic}[1]
\STATE \textbf{Input:} dataset $\mathcal{D}$, policy $\pi_\theta$, verifier, restructuring function $f_p$, rollout size $G$.
\FOR{each training step}
    \STATE Sample $\mathcal{D}_{\mathrm{batch}}$ and compute $\mathrm{Acc}(q,\pi_\theta)$ for each $q$ using on-policy rollouts.
    \STATE Partition active samples into hard, medium, and easy subsets.
    \STATE Restructure hard problems by simplification and medium problems by diversification.
    \STATE Evaluate restructured problems and compute $\Delta\mathrm{Acc}(q)$.
    \STATE Construct $B_{\mathrm{mix}}$ from original solving, restructured solving, and rewriting trajectories.
    \STATE Update $\pi_\theta$ with solving advantages and rewriting advantages.
\ENDFOR
\end{algorithmic}
\end{algorithm}

\subsection{Algorithm}
\label{subsec:algorithm}

Algorithm~\ref{alg:clpo} summarizes the full CLPO training loop. The procedure alternates between online difficulty diagnosis, adaptive restructuring, and restructuring-aware policy updates, so the curriculum is continually refreshed as the policy changes.

%% file: sec/4_experiment.tex
\section{Experiments}
\label{sec:experiments}


\subsection{Experimental Setup}
\label{subsec:setup}


\subsubsection{Base Models}
We use Qwen3-8B~\citep{yang2025qwen3} as the default backbone for our main experiments. To study the robustness across different model scales and families, we additionally evaluate Qwen3-4B, Qwen3.5-9B, Qwen3-30B-A3B, and Llama3.1-8B in Section~\ref{subsec:model_scales}. All variants are evaluated using identical decoding parameters.

\subsubsection{Baselines}
We compare CLPO against two categories of competitive fine-tuning methods.

\noindent\textbf{Supervised Fine-Tuning (SFT).}
These methods derive training signal from static or offline-generated 
supervision, without online environment feedback.
\textbf{RAFT}~\citep{dong2023raft} filters rollouts by reward ranking and 
fine-tunes on the retained outputs.
\textbf{Refinement-FT}~\citep{chen2024learning} constructs training 
examples from iteratively refined responses guided by natural language 
feedback.
\textbf{CITL-FT}~\citep{xi2024enhancing} jointly fine-tunes on both 
initial responses and critique-conditioned refinements.
\textbf{Critique-FT}~\citep{wang2025critique} directly supervises the 
model to produce critiques rather than final answers.

\noindent\textbf{Reinforcement Learning with Verifiable Rewards (RLVR).}
These methods use online rollout feedback as the primary training signals.
\textbf{GRPO}~\citep{shao2024deepseekmath} computes group-relative advantages over sampled rollouts under a scalar correctness reward.
\textbf{DAPO}~\citep{yu2025dapo} refines GRPO with clip-higher mechanisms and entropy bonuses to stabilize training under dynamic difficulty.
\textbf{LUFFY}~\citep{yan2025learning} blends off-policy expert trajectories with on-policy rollouts to regularize the learned policy.
\textbf{Critique-GRPO}~\citep{zhang2025critique} introduces natural-language critiques as an auxiliary reward signal within the GRPO framework, which we evaluate in both simple and \textsc{CoT} variants.
Unlike these methods, CLPO restructures problems online to control difficulty and jointly optimizes the restructuring behavior as a learnable policy component.

\begin{table*}[!tbp]
\centering
\normalsize
\setlength{\tabcolsep}{3.0pt}
\renewcommand{\arraystretch}{1.15}
\resizebox{\textwidth}{!}{%
\begin{tabular}{llccccccccc}
\toprule
\multirow{2}{*}{\textbf{Method}} & \multirow{2}{*}{\makecell{\textbf{Optimization} \\ \textbf{Policy}}}
& \multicolumn{5}{c}{\textbf{Math (In-Domain)}}
& \multicolumn{3}{c}{\textbf{General Reasoning (OOD)}}
& \multirow{2}{*}{\textbf{Avg}} \\
\cmidrule(lr){3-7}\cmidrule(lr){8-10}
& & \makecell{\textbf{MATH} \\ \textbf{500}} & \makecell{\textbf{Minerva} \\ \textbf{Math}} & \makecell{\textbf{Olympiad} \\ \textbf{Bench}} & \textbf{AMC23} & \textbf{AIME24} & \makecell{\textbf{MMLU} \\ \textbf{Pro}} & \makecell{\textbf{Theorem} \\ \textbf{QA}} & \makecell{\textbf{GPQA} \\ \textbf{Diamond}} & \\
\midrule
\multicolumn{11}{l}{\emph{All methods are finetuned on \textbf{Qwen3-8B}}} \\
\midrule
\multicolumn{11}{l}{\textcolor{gray}{\emph{Supervised Fine-Tuning (SFT)}}} \\
RAFT & \makecell[l]{Ranking-Based \\ Imitation} & 76.20 & 35.58 & 36.86 & 50.00 & 26.67 & 65.93 & 43.50 & 36.76 & 46.44 \\
Refinement-FT & \makecell[l]{Guided \\ Refinement} & 83.20 & 47.58 & 40.71 & 70.00 & 33.33 & 67.84 & 41.25 & 34.47 & 52.30 \\
Critique-FT & \makecell[l]{Learning to \\ Critique} & 79.00 & 35.23 & 39.64 & 67.50 & 33.33 & 63.16 & 46.00 & 34.84 & 49.84 \\
CITL-FT & \makecell[l]{Mixed-Data \\ SFT} & 76.40 & 37.20 & 38.57 & 62.50 & 30.00 & 66.13 & 44.25 & 36.36 & 48.93 \\
\midrule
\multicolumn{11}{l}{\textcolor{gray}{\emph{Reinforcement Learning with Verifiable Rewards (RLVR)}}} \\
GRPO & \makecell[l]{Group-Based \\ RL} & 89.20 & 51.47 & 57.40 & 82.50 & 43.33 & 69.86 & 54.75 & 47.80 & 62.04 \\
DAPO & \makecell[l]{Dynamic \\ Sampling} & \textbf{91.20} & 53.31 & 63.80 & 87.50 & 46.67 & 70.01 & 55.00 & 48.48 & 64.50 \\
LUFFY & \makecell[l]{Off-Policy \\ Imitation} & 89.40 & 52.94 & 58.80 & 85.00 & 40.00 & 70.34 & 58.25 & 49.49 & 63.03 \\
Critique-GRPO (Simple) & \makecell[l]{Critique-Driven \\ RL} & 89.40 & 52.57 & 60.20 & 87.50 & 40.00 & 70.13 & 59.00 & 48.63 & 63.43 \\
Critique-GRPO (CoT) & \makecell[l]{Critique-Driven \\ RL} & \textbf{91.20} & 61.50 & 63.80 & \textbf{90.00} & 46.67 & 70.98 & 59.50 & 50.50 & 66.77 \\
\textbf{CLPO (Ours)} & \makecell[l]{\textbf{Restructuring-Aware} \\ \textbf{RL}} & 90.80 & \textbf{72.79} & \textbf{77.50} & \textbf{90.00} & \textbf{50.00} & \textbf{73.12} & \textbf{67.75} & \textbf{56.06} & \textbf{72.25} \\
\bottomrule
\end{tabular}%
}
\caption{Main results on Qwen3-8B. We compare CLPO against competitive SFT and RLVR baselines across eight benchmarks. CLPO significantly outperforms all existing methods in both in-domain mathematical reasoning and out-of-distribution (OOD) general reasoning, demonstrating the effectiveness of self-evolving curriculum.}
\label{tab:qwen3_8b_main_results}
\end{table*}

\subsubsection{Benchmarks}
We evaluate our models across two categories of benchmarks.

\noindent\textbf{Mathematical Reasoning (In-Domain).}
We employ MATH-500~\citep{hendrycks2021measuring}, Minerva-Math~\citep{lewkowycz2022solving}, Olympiad Bench~\citep{he2024olympiadbench}, AMC23, and AIME2024~\citep{li2024numinamath}. These datasets cover a broad spectrum of difficulties, ranging from standard math competitions to advanced Olympiad levels.

\noindent\textbf{General Domain Reasoning (OOD).}
We utilize TheoremQA~\citep{chen2023theoremqa}, GPQA Diamond~\citep{rein2024gpqa}, and MMLU Pro~\citep{wang2024mmlu} as out-of-distribution benchmarks to assess generalization beyond the mathematical training domain.

All results are reported as Pass@1 accuracy. The primary aggregate metric is the average score across all eight benchmarks.

\subsubsection{Implementation Details}
For mathematical reasoning, we utilize the DAPO-Math-17k dataset~\cite{yu2025dapo}. For code generation (evaluated in the ablation studies in Section~\ref{subsec:ablations}), we adopt the code-domain subset of Eurus-2-RL-Data~\cite{cui2025process}. Unless otherwise specified, all compared methods use a rollout size of $G=4$ and binary verifier rewards ($\pm 1$) under a fixed RLVR optimization budget. All experiments are conducted on a node equipped with 8$\times$ NVIDIA H20 GPUs. Full hyperparameter configurations and CLPO-specific thresholds ($\tau_{\text{hard}}$, $\tau_{\text{easy}}$, $\lambda$) are detailed in Appendix~\ref{app:implementation_details}.

\subsection{Main Results}
\label{subsec:main_results}

Table~\ref{tab:qwen3_8b_main_results} summarizes the performance of CLPO on Qwen3-8B. CLPO achieves a state-of-the-art average score of 72.25\%, outperforming the strongest RLVR baseline, Critique-GRPO (CoT), by 5.48 points and DAPO by 7.75 points. The performance gain is particularly significant on challenging math benchmarks: CLPO sets new competitive records on Minerva Math (+11.29\%), Olympiad Bench (+13.7\%), and AIME24 (+3.33\%), while maintaining top-tier performance on AMC23 and MATH 500. This suggests that restructuring-aware optimization successfully converts difficulty signals into high-quality training curriculum rather than redundant rollouts.

The superiority of CLPO extends to OOD reasoning. Across MMLU Pro, TheoremQA, and GPQA Diamond, CLPO consistently achieves the highest scores among all baselines. This generalization suggests that CLPO fosters robust reasoning capabilities rather than dataset-specific overfitting. Compared to SFT-style methods, CLPO exhibits a substantial margin (at least 19.95\% average gain), underscoring the necessity of verifiable online feedback for complex reasoning tasks.



\subsection{Training Dynamics and Ablation Studies}
\label{subsec:ablations}

\begin{figure}[!tbp]
    \centering
    \makebox[\columnwidth][c]{%
    \begin{minipage}{1.0\columnwidth}
        \centering
        \begin{minipage}[t]{0.49\linewidth}
            \centering
            \includegraphics[width=\linewidth]{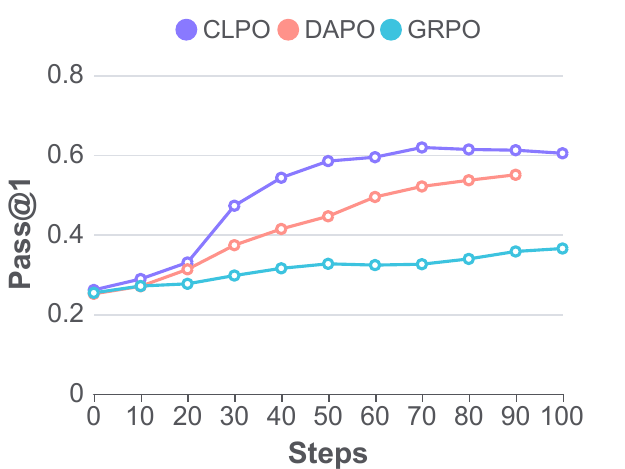}\\[-0.3em]
            {\small (a) Code training dynamics}
        \end{minipage}\hfill
        \begin{minipage}[t]{0.49\linewidth}
            \centering
            \includegraphics[width=\linewidth]{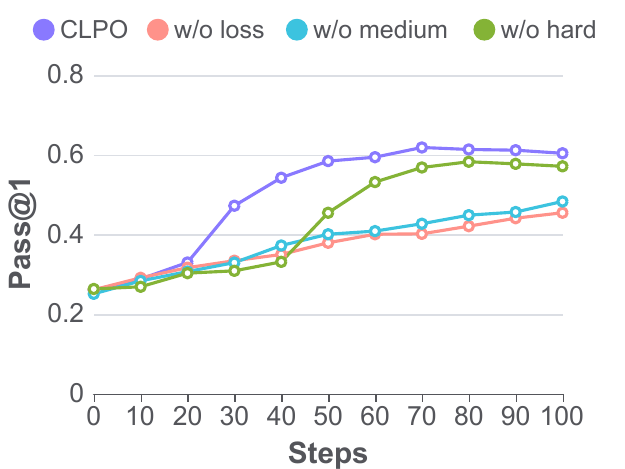}\\[-0.3em]
            {\small (b) Code ablation study}
        \end{minipage}

        \vspace{0.25em}
        \begin{minipage}[t]{0.49\linewidth}
            \centering
            \includegraphics[width=\linewidth]{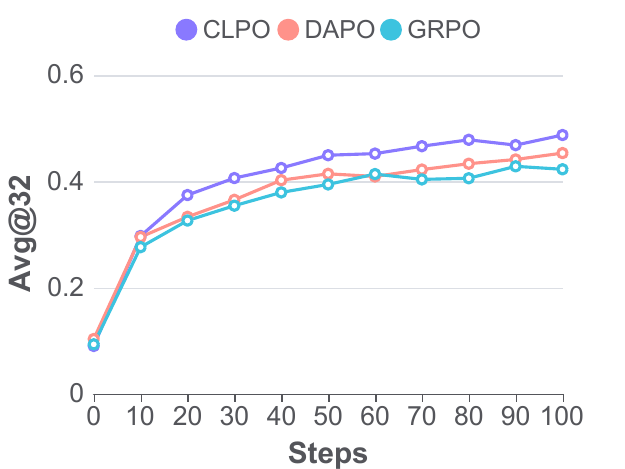}\\[-0.3em]
            {\small (c) Math training dynamics}
        \end{minipage}\hfill
        \begin{minipage}[t]{0.49\linewidth}
            \centering
            \includegraphics[width=\linewidth]{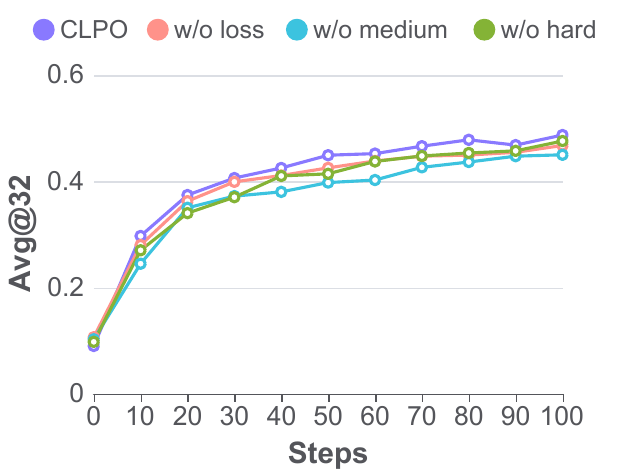}\\[-0.3em]
            {\small (d) Math ablation study}
        \end{minipage}
    \end{minipage}%
    }
    \caption{Training dynamics and ablation results on code-oriented and mathematical reasoning tasks. The left column compares CLPO with GRPO and DAPO, while the right column evaluates key CLPO variants.}
    \label{fig:exp2_learning_ablation}
\end{figure}

\noindent\textbf{Training dynamics.}
Figure~\ref{fig:exp2_learning_ablation} (left) illustrates the superiority of CLPO in both training efficiency and endpoint performance. On the code task, CLPO achieves 60.35\% Pass@1, substantially outperforming DAPO (45.43\%) and GRPO (36.45\%). Similar trends are observed in the math domain (48.70\% vs. 45.30\% for DAPO), validating CLPO's robustness across diverse reasoning tasks. 

\noindent\textbf{Component ablation.}
Figure~\ref{fig:exp2_learning_ablation} (right) investigates the contribution of each module. Removing the restructuring loss (\textit{w/o loss}) causes the most significant performance drop (60.35\% $\rightarrow$ 45.43\% on code), demonstrating that restructuring should be optimized as a learnable skill rather than treated as static data augmentation. On the math task, removing medium-difficulty diversification (\textit{w/o medium}) results in the lowest accuracy (44.98\%), while removing hard-problem simplification (\textit{w/o hard}) also degrades performance. This confirms their complementary roles: diversification strengthens robustness at the model's capability frontier, while simplification recovers learning signals from otherwise uninformative instances.

\subsection{Training Efficiency under Fixed Budgets}
\label{subsec:efficiency}
As shown in Table~\ref{tab:time_budget_comparison}, CLPO demonstrates superior training efficiency under fixed wall-clock budgets. Within a 10-hour window, CLPO reaches an average of 38.02 points, a level that baselines fail to achieve even with higher update counts. This advantage persists at the 20-hour mark (43.09 avg.), proving that CLPO extracts more informative learning signals per unit of compute time.

Interestingly, the step counts highlight that performance is not strictly tied to update frequency. CLPO achieves its lead with an intermediate number of steps (43 and 88), positioned between the high-throughput GRPO and the lower-throughput DAPO. This validates that the computational investment in online problem restructuring yields a superior "return on investment" compared to baseline RLVR methods, turning available time into more robust reasoning capabilities.



\begin{table}[!tbp]
\centering
\small
\setlength{\tabcolsep}{3pt}
\renewcommand{\arraystretch}{1.15}
\resizebox{\columnwidth}{!}{%
\begin{tabular}{lcccccc}
\toprule
\textbf{Method} & \makecell{\textbf{Time} \\ \textbf{Budget}} & \textbf{Steps} & \textbf{AIME24} & \textbf{AIME25} & \textbf{AIME26} & \textbf{Avg.} \\
\midrule
\multirow{2}{*}{GRPO}
& 10h & 53  & 40.73 & 35.21 & 34.06 & 36.67 \\
& 20h & 107 & 43.96 & 41.77 & 40.00 & 41.91 \\
\midrule
\multirow{2}{*}{DAPO}
& 10h & 24 & 38.23 & 36.77 & 34.27 & 36.42 \\
& 20h & 52 & 41.98 & 42.40 & \textbf{40.73} & 41.70 \\
\midrule
\multirow{2}{*}{\textbf{CLPO}}
& 10h & 43 & \textbf{41.35} & \textbf{37.08} & \textbf{35.63} & \textbf{38.02} \\
& 20h & 88 & \textbf{45.94} & \textbf{42.92} & 40.42 & \textbf{43.09} \\
\bottomrule
\end{tabular}%
}
\caption{Time-budget comparison on AIME benchmarks. CLPO achieves the best average performance under both 10-hour and 20-hour budgets, showing stronger training efficiency than GRPO and DAPO.}
\label{tab:time_budget_comparison}
\end{table}

\begin{figure}[!tbp]
    \centering
    \makebox[\columnwidth][c]{%
    \begin{minipage}{1.0\columnwidth}
        \centering
        \begin{minipage}[t]{0.49\linewidth}
            \centering
            \includegraphics[width=\linewidth]{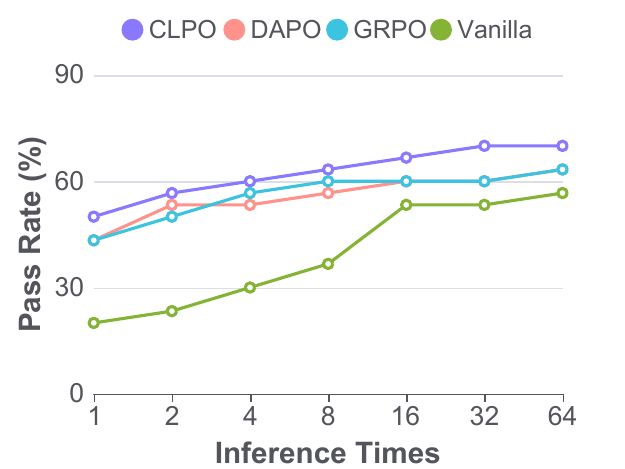}\\[-0.35em]
            {\scriptsize (a) AIME 2024}
        \end{minipage}\hfill
        \begin{minipage}[t]{0.49\linewidth}
            \centering
            \includegraphics[width=\linewidth]{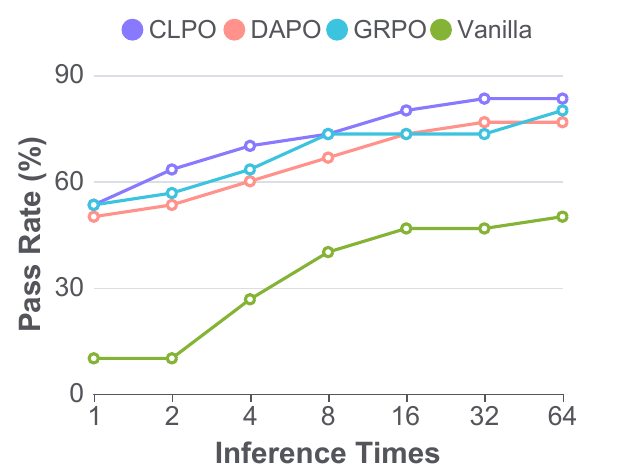}\\[-0.35em]
            {\scriptsize (b) AIME 2025}
        \end{minipage}

        \vspace{0.1em}
        \begin{minipage}[t]{0.49\linewidth}
            \centering
            \includegraphics[width=\linewidth]{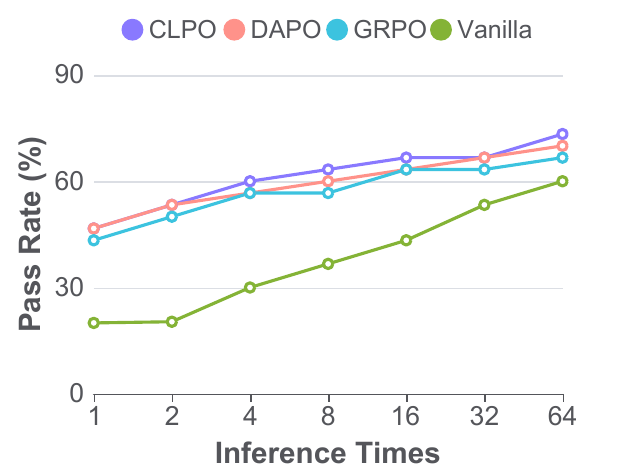}\\[-0.35em]
            {\scriptsize (c) AIME 2026}
        \end{minipage}
    \end{minipage}%
    }
    \caption{Test-time scaling results of Qwen3-8B on AIME 2024, AIME 2025, and AIME 2026.}
    \label{fig:test_time_scaling}
\end{figure}

\subsection{Test-Time Scaling}
\label{subsec:scaling}



Figure~\ref{fig:test_time_scaling} demonstrates that CLPO scales effectively with test-time compute, reaching 70.00\%, 83.33\%, and 73.33\% accuracy on AIME 2024--2026 at 64 samples. Notably, on AIME 2025, CLPO achieves a 30.0-point gain over its Pass@1 baseline, consistently outperforming GRPO and DAPO. These results indicate that CLPO enhances the policy's \textit{searchability} by producing a higher density of correct reasoning paths within the solution distribution. 

\subsection{Model-Scale Analysis}
\label{subsec:model_scales}

\begin{figure}[H]
    \centering
    \includegraphics[width=0.76\columnwidth]{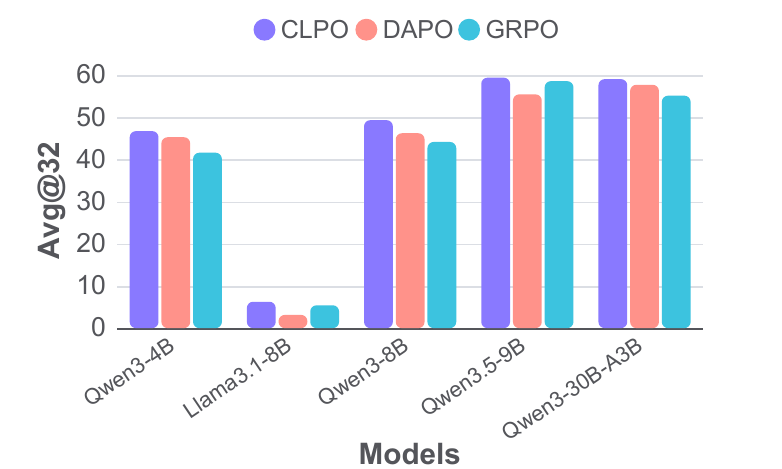}
    \caption{Model-scale comparison across representative base models on AIME24.}
    \label{fig:model_scales}
\end{figure}

Figure~\ref{fig:model_scales} investigates the generalizability of CLPO across various base-model scales and architectures on AIME24. CLPO consistently achieves the highest performance on all tested backbones, outperforming both DAPO and GRPO across: Qwen3-4B (46.7\% vs. 45.3\%/41.6\%), Llama3.1-8B (6.3\% vs. 3.1\%/5.4\%), Qwen3-8B (49.3\% vs. 46.2\%/44.1\%), Qwen3.5-9B (59.4\% vs. 55.4\%/58.5\%), and Qwen3-30B-A3B (59.0\% vs. 57.7\%/55.1\%).

The model-scale comparison suggests that CLPO complements scaling rather than specializing to one checkpoint. Larger models generally have stronger initial reasoning ability, but CLPO can still convert intermediate online signals into useful training examples.


%% file: sec/5_conclusion.tex
\section{Conclusion}
\label{sec:conclusion}

We present CLPO, a self-evolving curriculum framework for RLVR that uses rollout accuracy to diagnose current competence, restructure problems around the productive learning zone, and optimize restructuring behavior through downstream accuracy improvement. By turning the model from a passive solver into an active curriculum constructor, CLPO achieves consistent gains over strong RLVR baselines.

%% file: sec/F_llm_usage.tex
\section{Usage of Large Language Models}
\label{app:llm_usage}

During the preparation of this manuscript, we utilized Large Language Models (LLMs) as a writing assistant. The usage of LLMs was strictly limited to improving the fluency, clarity, and grammatical correctness of the language, such as rephrasing sentences or correcting grammatical errors. LLMs were not involved in the core research ideation, experimental design, analysis of results, or the formulation of conclusions presented in this paper.

%% file: sec/A_implementation_details.tex
\section{Implementation Details}
\label{app:implementation_details}

\begin{table}[t]
\centering
\small
\begin{tabular}{ll}
\toprule
\textbf{Item} & \textbf{Description} \\
\midrule
Dataset & Eurus-2-RL-Data \\
Domain & Code \\
Source & APPS, CodeContests, TACO, Codeforces \\
Scale & 25,276 train / 1,024 valid \\
\bottomrule
\end{tabular}
\caption{Code training data.}
\label{tab:code_training_data}
\end{table}

\begin{table*}[t]
\centering
\small
\label{tab:main_hyperparams}
\begin{tabular}{p{0.23\textwidth} p{0.20\textwidth} p{0.47\textwidth}}
\toprule
\textbf{Name} & \textbf{Value} & \textbf{Description} \\
\midrule
\multicolumn{3}{l}{\textit{Training}} \\
Base Model & Qwen3-8B & The base model used in experiments. \\
Dataset & DAPO-Math-17k & The dataset used for training. \\
training\_steps & 200 & Total number of training steps. \\
Optimizer & AdamW & The optimizer used. \\
lr & 1e-6 (Constant) & Learning rate. \\
batch\_size & 64 & Global batch size during training. \\
n\_rollouts ($G$) & 4 & Number of rollouts per prompt. \\
rewards & 1 or -1 & Scalar rewards for correct/incorrect responses. \\
kl\_loss\_coef ($\beta$) & 0.001 & Coefficient for KL divergence loss. \\
grad\_clip & 1.0 & Gradient clipping threshold. \\
train\_temp & 1.0 & Sampling temperature during training rollout. \\
top\_p & 1.0 & Top-p sampling parameter during training rollout. \\
max\_response\_length & 8192 & Maximum length of generated responses. \\
\midrule
\multicolumn{3}{l}{\textit{Evaluation}} \\
val\_temp & 1.0 & Sampling temperature during evaluation. \\
\midrule
\multicolumn{3}{l}{\textit{Environment}} \\
Hardware & 8 $\times$ NVIDIA H20 & Hardware used for experiments. \\
Software & verl~\citep{sheng2025hybridflow} & Software frameworks used. \\
\bottomrule
\end{tabular}
\caption{CLPO-specific hyperparameters.}
\label{tab:main_hyperparams}
\end{table*}

\begin{table*}[t]
\centering
\small

\begin{tabular}{p{0.30\textwidth} p{0.16\textwidth} p{0.44\textwidth}}
\toprule
\textbf{Name} & \textbf{Value} & \textbf{Description} \\
\midrule
hard threshold ($\tau_{\text{hard}}$) & 0.25 & With $G=4$, problems with one successful rollout are treated as hard. \\
easy threshold ($\tau_{\text{easy}}$) & 0.75 & Problems with three or more successful rollouts are treated as easy. \\
rewrite loss weight ($\lambda$) & 0.1 & Weight for the restructuring-aware rewriting objective. \\
active-region filter & $(0,1)$ & Original and restructured problems are kept when rollout accuracy is non-degenerate. \\
\bottomrule
\end{tabular}
\vspace{1em}
\caption{Main hyperparameters for training, evaluation, and environment.}
\label{tab:clpo_hyperparams}
\end{table*}

This section summarizes the implementation details needed to reproduce our experiments. All training and evaluation runs use the verl framework~\citep{sheng2025hybridflow} with vLLM~\citep{kwon2023efficient} for fast rollout. Table~\ref{tab:main_hyperparams} reports the common training, evaluation, and hardware settings, while Table~\ref{tab:clpo_hyperparams} lists the CLPO-specific hyperparameters.

The main comparison uses Qwen3-8B as the base model. We additionally evaluate Qwen3-4B, Qwen3.5-9B, Qwen3-30B-A3B, and Llama3.1-8B-Instruct in the model-scale analysis. The benchmark suite covers mathematical reasoning and general reasoning, including MATH-500, MinervaMath, Olympiad Bench, AMC23, AIME2024/2025/2026, MMLU Pro, TheoremQA, and GPQA Diamond. For fixed-budget studies, we run 10-hour and 20-hour AIME training budgets and report both performance and optimization steps.

For the code setting, we use only the code-domain subset of Eurus-2-RL-Data\footnote{https://huggingface.co/datasets/PRIME-RL/Eurus-2-RL-Data}, which is curated from APPS, CodeContests, TACO, and Codeforces. We follow its official train/validation split, containing 25,276 training examples and 1,024 validation examples, as summarized in Table~\ref{tab:code_training_data}.

\begin{figure*}[t]
    \centering
    \includegraphics[width=0.95\textwidth]{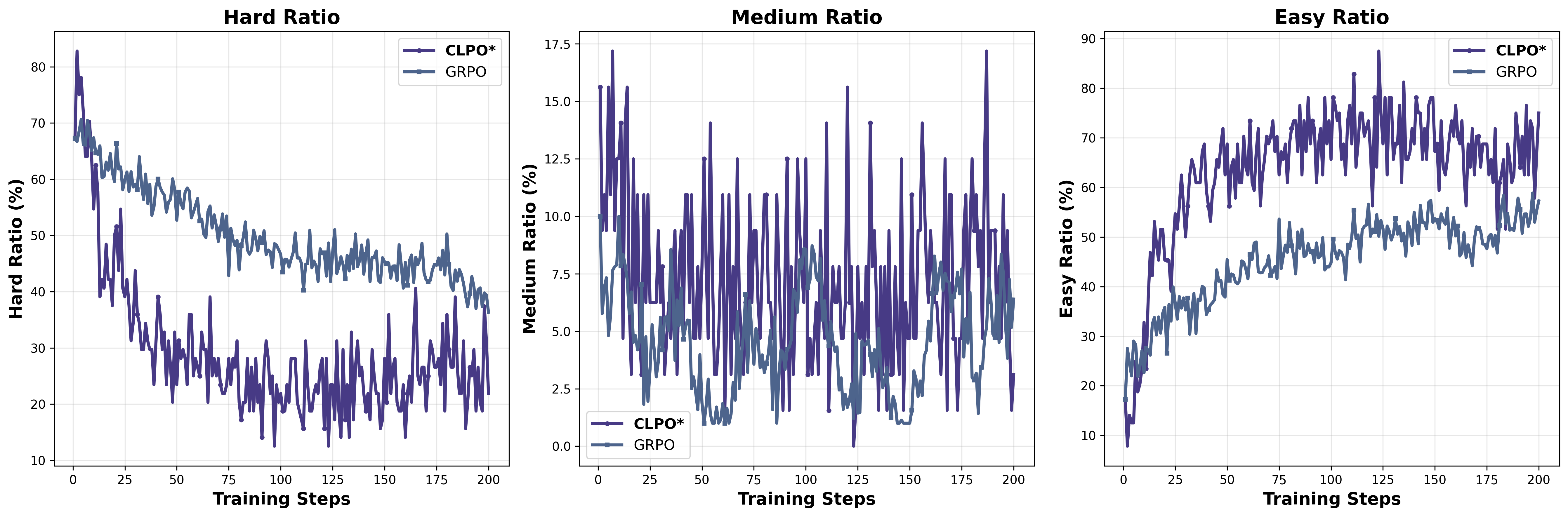}
    \caption{Evolution of problem difficulty distribution during training for CLPO and GRPO over the first 200 training steps.}
    \label{fig:curriculum_dynamics}
\end{figure*}

%% file: sec/C_curriculum_dynamics.tex
\section{Visualization of Curriculum Dynamics}
\label{app:curriculum_dynamics}

Figure~\ref{fig:curriculum_dynamics} visualizes the problem difficulty distribution in dynamically constructed training batches. Compared with GRPO, CLPO moves hard problems out of the unproductive region faster and maintains a larger supply of medium-difficulty examples, which are most useful for RLVR updates.

\paragraph{Hard Ratio.} CLPO shows a sharper early decline in hard problems than GRPO, indicating that simplification helps recover useful gradients from otherwise difficult instances.

\paragraph{Medium Ratio.} CLPO maintains a higher medium-problem ratio, showing that restructuring continually replenishes the active learning region near the model's capability frontier.

\paragraph{Easy Ratio.} CLPO also grows the easy-problem ratio faster, reflecting more efficient conversion from learnable examples to mastered examples.

Together, these dynamics support the claim that CLPO improves training not only by selecting samples, but also by restructuring them into more informative curriculum instances.

%% file: sec/E_case_studies.tex
\begin{figure*}[h]
    \centering
    \includegraphics[width=0.86\textwidth, height=0.82\textheight, keepaspectratio]{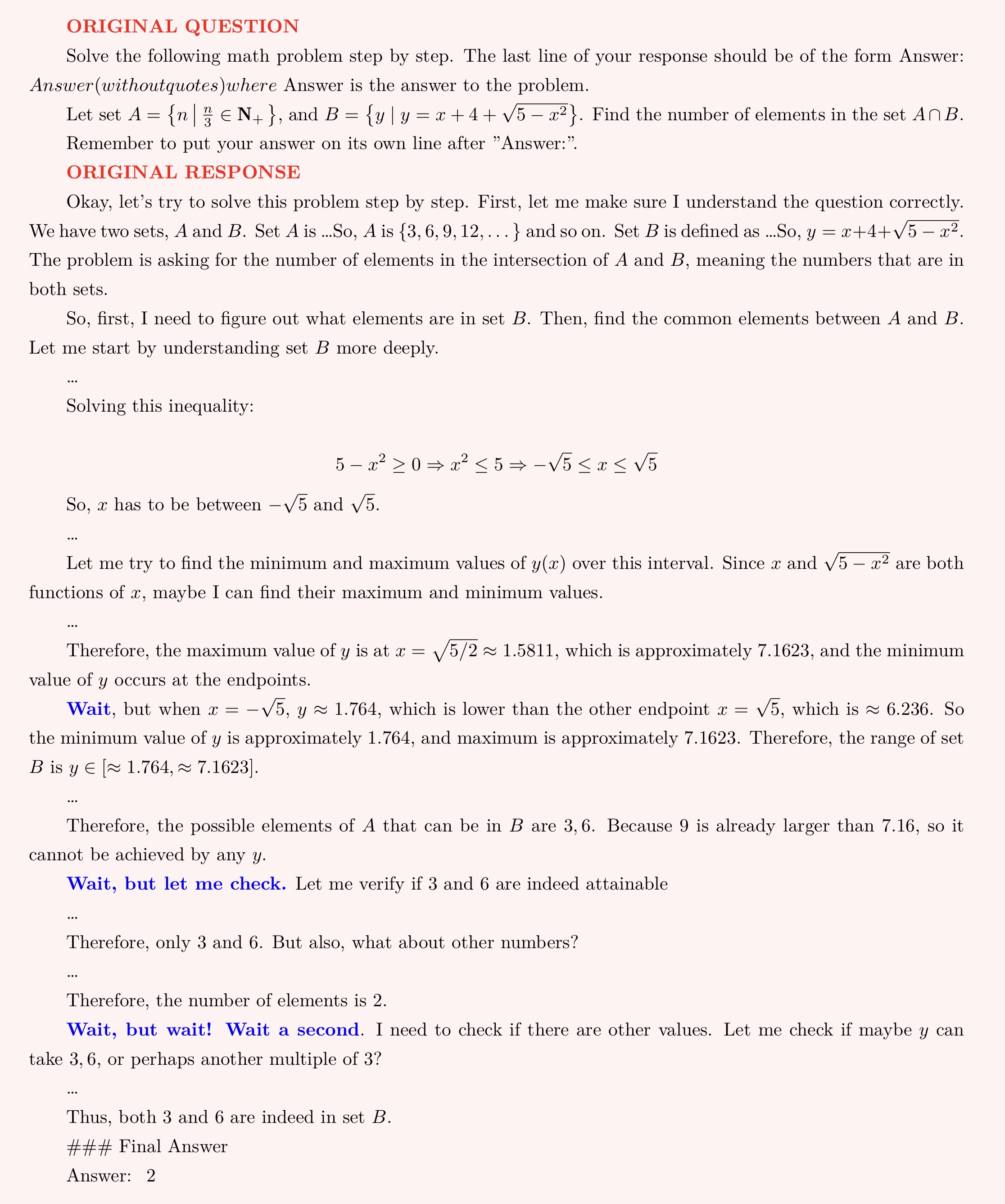}
    \caption{The original medium-difficulty problem presented in a symbolic style and the corresponding hesitant response generated by the Qwen3-8B model.}
    \label{fig:case_original}
\end{figure*}

\begin{figure*}[h]
    \centering
    \includegraphics[width=0.86\textwidth, height=0.82\textheight, keepaspectratio]{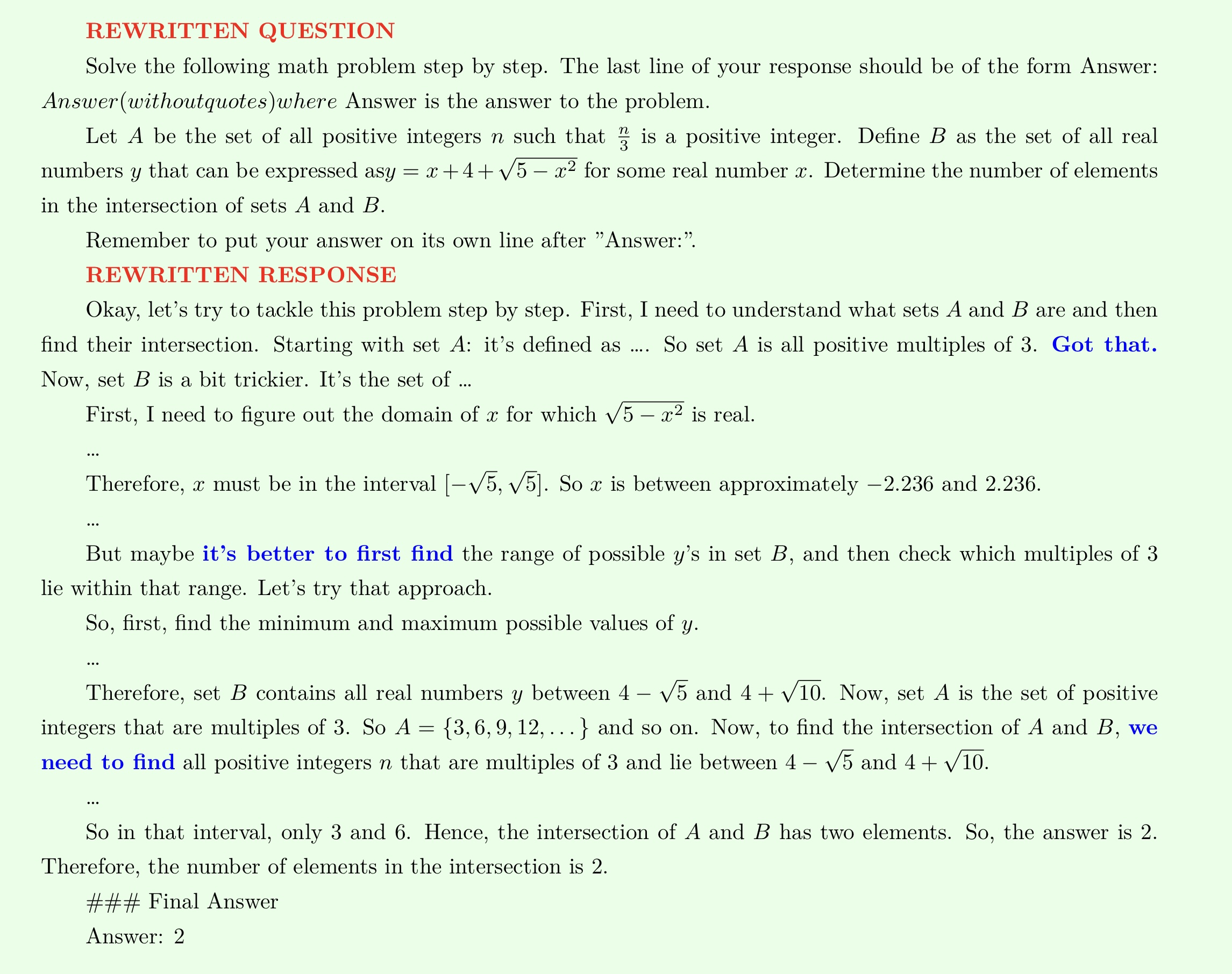}
    \caption{The restructured version of the problem, elaborated into natural language by CLPO, and the resulting confident, linear reasoning path from the same model.}
    \label{fig:case_restructured}
\end{figure*}

\section{Case Studies}
\label{app:case_studies}
To provide a concrete illustration of our Adaptive Problem Restructuring (APR) mechanism in action, this section presents a case study on a medium-difficulty problem. We showcase the original problem and the model's response, followed by the restructured version generated by CLPO and the corresponding improvement in the model's reasoning process.

\subsection{Case Study: Diversification of a Medium-Difficulty Problem}

Figure~\ref{fig:case_original} displays a medium-difficulty problem defined using a compact, symbolic notation typical in mathematical texts. The response generated by the base Qwen3-8B model, while ultimately correct, exhibits a hesitant and circuitous reasoning path. The model frequently expresses self-doubt (e.g., "Wait, but let me check," "Wait a second") and performs multiple, redundant verification steps. This indicates that while the model possesses the necessary knowledge, the symbolic formulation of the problem introduces ambiguity, leading to a less confident and inefficient solution process.

Figure~\ref{fig:case_restructured} shows the same problem after being processed by CLPO's diversification mechanism. The algorithm restructured the problem by elaborating the compact, symbolic set-builder notation into explicit, full-sentence natural language descriptions. This seemingly minor change had a profound impact on the model's response. The new reasoning path is linear, structured, and confident, proceeding logically from step to step without the self-doubt and redundant checks seen previously. 

This case study vividly demonstrates the effectiveness of our APR mechanism. By transforming a problem into a linguistic format that is more aligned with the model's pre-training data, CLPO is able to reduce ambiguity and significantly improve the robustness and efficiency of the reasoning process itself, not just the final answer.

\section{Prompts for Adaptive Problem Restructuring}
\label{app:prompts}
See our prompt templates for adaptive problem restructuring in \textbf{Prompt for Hard Prompt Simplification} and \textbf{Prompt for Question Diversification}.

%% file: sec/D_prompts.tex
\clearpage

\begin{appendixbox}{Prompt for Hard Prompt Simplification}
You are a master-level question rewriting expert. Your mission is TOP-SECRET: rewrite the given math question so that it is simpler, clearer, and more direct, while maintaining the EXACT same task, constraints, solution method, and CORRECT ANSWER. DO NOT solve the problem or make any assumptions about its solution.

\textbf{CRITICAL RULES FOR OUTPUT:
}

\textbf{- Do NOT include any reasoning, parsing, or explanation of the problem (e.g., avoid phrases like "Let me simplify this question" or "This problem can be rewritten as...").
}

\textbf{- Rewrite the question to make it SIMPLER and CLEARER for the reader, while preserving the EXACT task, math content, and structure.
}

\textbf{- Keep the rewritten question STRICTLY UNDER 400 tokens, including all characters, symbols, and spaces.
}

\textbf{- Remove redundancy and completely clarify definitions, if needed. For example, if there are implied constraints, make them explicit.
}

\textbf{- Preserve ALL formatting or directive rules exactly as in the original question (e.g., "write the answer in a box", "solve step by step").
}

\textbf{- The rewritten question must lead to the SAME correct answer as the original answer. DO NOT explicitly, implicitly, or indirectly REVEAL or SUGGEST the answer.
}

HOW TO SIMPLIFY:

1) Eliminate redundant or overly verbose phrasing while keeping the exact math concepts, symbols, and structure.

2) Avoid complex clauses by breaking them into shorter, simpler sentences, but do NOT remove or alter necessary constraints.

3) Include IMPLICIT constraints explicitly if they improve clarity (e.g., "x > 0" if it is implied implicitly).

4) DO NOT add any background information, new context, or reasoning unrelated to the question.

\end{appendixbox}
\medskip

\begin{appendixbox}{Prompt for Question Diversification}
You are a master-level question rewriting expert. Your mission is TOP-SECRET:
rewrite the given math question into a diverse but semantically equivalent
version, while maintaining the EXACT same task, constraints, solution method,
and CORRECT ANSWER. DO NOT solve the problem or make any assumptions about its
solution.

\textbf{CRITICAL RULES FOR OUTPUT: }

\textbf{- Rewrite the question to DIVERSIFY its expression but PRESERVE its
exact meaning and constraints. }

\textbf{- Keep the rewritten question STRICTLY UNDER 400 tokens, including all
characters, symbols, and spaces. }

\textbf{- DO NOT change any math content (variables, relationships, solution
method, etc.). }

\textbf{- Avoid repetitive sentence structures or template-like phrasing by
using varied grammar structure, synonyms, or logical order. }

\textbf{- Preserve ALL formatting or directive rules exactly as in the original
question (e.g., "write the answer in a box", "solve step by step"). }

\textbf{- The rewritten question must lead to the SAME correct answer as the
original: {ANSWER}. DO NOT explicitly, implicitly, or indirectly REVEAL or
SUGGEST the answer. }

HOW TO DIVERSIFY:

1)  Rephrase clauses logically (e.g., change phrasing while maintaining meaning).

2)  Rearrange sentence structure without altering mathematical intent.

3)  Use synonyms or alternative phrasing to express the same logic and tasks.

4) DO NOT add or remove constraints, context, or instructions.

\end{appendixbox}